\documentclass[letterpaper]{article} 
\usepackage[preprint]{aaai2027}  
\usepackage[hyphens]{url}  
\usepackage{graphicx} 
\usepackage{natbib}  
\usepackage{caption} 
\usepackage{algorithm}
\usepackage{algorithmic}

\usepackage{newfloat}
\usepackage{listings}
\DeclareCaptionStyle{ruled}{labelfont=normalfont,labelsep=colon,strut=off} 
\floatstyle{ruled}
\newfloat{listing}{tb}{lst}{}
\floatname{listing}{Listing}

\usepackage{booktabs}
\usepackage{graphicx}
\usepackage{subcaption}
\usepackage{multirow}
\usepackage{array}
\usepackage{amsmath}
\usepackage{amssymb}
\title{Shift-Aware Calibration for Fine-Tuned CLIP: Leveraging Image-Text Alignment}
\author{
    Song-Lin Lv\textsuperscript{\rm 1,\rm 2},
    Yu-Yang Chen\textsuperscript{\rm 1,\rm 2},
    Zhi Zhou\textsuperscript{\rm 2},
    Lan-Zhe Guo\textsuperscript{\rm 1,\rm 2}\corresponding
}
\affiliations{
    \textsuperscript{\rm 1}School of Intelligence Science and Technology, Nanjing University, China\\
    \textsuperscript{\rm 2}National Key Laboratory for Novel Software Technology, Nanjing University, China\\
    guolz@lamda.nju.edu.cn
}

\begin{document}

\maketitle

\begin{abstract}
Vision-language models (VLMs), such as CLIP, adapt effectively to downstream tasks through prompt tuning, but fine-tuning can misalign predictive confidence and accuracy, particularly on unseen classes. Existing VLM-specific calibration methods mainly rely on textual features of train classes, limiting their applicability across class and distribution shifts. We propose \textbf{Shift-Aware Calibration (SAC)}, a training-free, sample-wise method that uses the discrepancy between the output logits of the original and fine-tuned CLIP as a calibration signal. We explicitly call this quantity \emph{logit shift}, since it is measured in prediction space rather than between hidden representations. SAC maps this logit shift to a positive scaling factor that adjusts confidence while preserving the predicted class. Across 11 datasets and five fine-tuning methods, SAC improves calibration on train and unseen classes and under cross-dataset and domain-generalization evaluation. Mutual Information (MI) analysis shows a statistical dependency between the logit-shift signal and ECE across methods and datasets, providing empirical evidence that the signal used by SAC is informative for calibration and supporting the effectiveness of its shift-aware design.
\end{abstract}

\section{Introduction}
Vision-language models, such as CLIP~\cite{CLIP}, pre-trained on vast web-scale text-image datasets, have demonstrated impressive zero-shot capabilities and image-text alignment in downstream image classification tasks \cite{imagenet,eurosat}. Concurrently, various prompt learning methods for VLMs have been proposed to further enhance the performance of VLMs on specific tasks by leveraging a small amount of labeled data~\cite{maple,promptsrc,coop}. Given CLIP’s strong zero-shot adaptability, the open-vocabulary setting has become a standard for evaluating the performance of fine-tuned VLMs, where prompts are trained on a subset of classes and evaluated on both train and unseen classes~\cite{lee2023read,tan2024compound}.

Unfortunately, fine-tuned VLMs regularly overfit to train classes, forgetting the well-calibrated predictions and image-text alignment achieved during pre-training~\cite{cocoop,coop}. 
For unseen classes, they often produce semantically unbalanced representations, leading to image-text misalignment and a significant discrepancy between confidence scores and actual accuracy~\cite{modern,remodern}.
Existing calibration methods~\cite{ATS,Oncalibration,bin1} typically fit or analyze statistics from the training distribution, which limits their use on unseen classes. Distance-Aware Calibration (DAC)~\cite{DAC}, a VLM-specific method, derives class-wise weights from textual representations and is designed primarily for unseen classes within the same dataset; it does not directly model sample-level visual shifts in cross-dataset or domain-generalization evaluation.

Prior work~\cite{remodern,closelook,promptsrc} suggests that the original CLIP benefits from well-aligned image--text representations and often exhibits favorable calibration. Motivated by this observation, we analyze fine-tuning dynamics and find that changes in calibration are associated with a mismatch between the shift in model outputs and the corresponding change in accuracy. Because our measurable quantity is computed from logits rather than hidden features, we refer to it as \textit{logit shift}; it serves as an observable proxy for the prediction-space effects of fine-tuning.

Based on this analysis, we develop \textbf{Shift-Aware Calibration (SAC)}. SAC uses the discrepancy between the logits of the original CLIP (the ``calibration anchor'') and the fine-tuned CLIP to compute a sample-wise scaling coefficient. The coefficient is positive, so rescaling changes confidence without changing the predicted class.
 
Due to its design tailored for CLIP with strong multimodal alignment, SAC can be applied to any set of classes and any evaluation settings, such as \textbf{open-vocabulary, cross-dataset, and domain-generalization} across 11 datasets. Using default parameters, SAC consistently delivers calibration improvements across all five fine-tuning methods and achieves the best performance in confidence calibration, outperforming the current best calibration method MIR~\cite{bin3} and SOTA VLM-specific calibration method DAC~\cite{DAC}. Notably, SAC demonstrates robust performance across various potential variations, including training sample shots, model backbones, datasets, fine-tuning methods, and evaluation settings, highlighting its insensitivity to parameters.

In summary, the main contributions of this paper include:
\begin{itemize} 
\item We identify that the key driving factor of confidence imbalance of fine-tuned CLIP is the mismatch between the degree of logit shift and the corresponding changes in accuracy. This finding provides a reliable reference for future research aimed at addressing confidence calibration issues in fine-tuned models.

\item Based on the analysis, we propose a novel method, Shift-Aware Calibration (SAC). As a post-hoc and fine-grained calibration approach, SAC achieves sample-level confidence calibration without compromising the model's output accuracy.

\item  Benefiting from the design of using the original CLIP as a ``calibration anchor'', SAC overcomes the limitations of existing methods and extends the experimental scope to all VLM evaluation settings, including open-vocabulary, cross-dataset, and domain-generalization.

\item The experimental results on 11 datasets, testing the calibration of 5 fine-tuning methods, show that SAC outperforms existing calibration methods. SAC demonstrates strong robustness and parameter insensitivity.
\end{itemize}

\section{Related Works}
\textbf{Prompt Learning in Vision Language Models.} Due to the large parameter size of VLMs and the limited availability of training data for downstream tasks, it is impractical to fine-tune all parameters of the VLMs to adapt them to these tasks. To improve the generalization of the learnable language prompt~\cite{coop, iclr1,eccv2}, CoCoOp~\cite{cocoop} generates a vision-conditional prompt by fusing the image feature and the learnable language prompts. KgCoOp~\cite{kgcoop} and ProGrad~\cite{prograd} are other prompt-based methods for VLMs. MaPLe~\cite{maple} and PromptSRC~\cite{promptsrc} conduct the visual-textual prompt learning by jointly conducting the prompt learning on the vision and text encoders. To address higher confidence bias for unseen classes caused by fine-tuning, the proposed method leverages the original CLIP to correct the confidence of fine-tuned CLIP.

\textbf{Confidence Calibration.} Confidence calibration aims to align the confidence scores predicted by models with their actual performance. A common strategy for achieving this is to apply calibration techniques after model training. These techniques can be broadly divided into two categories: scaling-based methods~\cite{modern,proximity,ts} and bin-based methods~\cite{bin3,bin1,bin2}. Distance-Aware Calibration~\cite{DAC} estimates class-wise scaling weights for unseen classes from textual representations, but its original formulation is limited when the target class vocabulary or visual distribution changes substantially. As a newly proposed method, SAC enables effective sample-level calibration across common experimental scenarios in VLM research, with a detailed comparison to existing methods provided in Table~\ref{tab:method-comparison}. In addition, some test-time adaptation~\cite{c-tpt, o-tpt} and fine-tuning~\cite{oh2024towards} methods have also discussed confidence; however, since these methods affect the accuracy of the original model and differ substantially from calibration approaches, we do not consider them here.

\begin{table*}[!t]
    \centering
    \begin{tabular}{lcccc}
    \toprule
        Method & Train & Required Information & CG & Applicable Scenarios \\ \midrule
        Bin Methods & No & Logits and ground-truth labels & Dataset & train classes only  \\ 
        Scaling Methods & Yes & Logits and ground-truth labels & Dataset & train classes only \\
        DAC & No & Textual representations of all classes & Class &  unseen classes in the same dataset \\
        SAC (Ours) & No & Original and fine-tuned logits on testing data & Sample & train, unseen classes, cross-dataset... \\
        \bottomrule
    \end{tabular}
    \caption{Comparison of existing confidence calibration methods in terms of required information and applicable scenarios. ``CG'' represents the Calibration Granularity.}
    \label{tab:method-comparison}
\end{table*}

\textbf{Calibration Evaluation Metrics.} A model is perfectly calibrated if \( \mathbb{P}(\hat{y} = y|\hat{p} = p) = p\) for all \(p\) in \([0,1]\),
 where \(y\) is the actual label, \(\hat{y}\) the prediction, and \(\hat{p}\) the confidence score. To assess model calibration, we typically use the Expected Calibration Error (ECE)~\cite{modern} , lower values indicating better calibration. ECE groups predictions into \(M\) interval bins (each of size \(1/M\) ) and calculate the accuracy of each bin. The ECE is defined as the difference between the accuracy and confidence of all bins, which can be calculated as: \(\text{ECE} = \sum_{m=1}^M\frac{|B_m|}{n}|acc(B_m) - conf(B_m)|\). We also report three additional metrics: Maximum Calibration Error (MCE)~\cite{modern}, Adaptive Calibration Error (ACE)~\cite{ace} and Proximity Informed Expected Calibration Error (PIECE)~\cite{proximity}.

\section{Preliminary}
\textbf{Zero-shot Classification of Pre-trained CLIP.} CLIP aligns visual and textual data in a shared embedding space using an image encoder \(f\) and a text encoder \(g\). Given a caption T, the text encoder produces features \(g(E_w(T))\), where \(E_w\) is the word embedding layer. In zero-shot classification, CLIP uses prompts of the form \(a\ photo\ of\ a\ [CLASS_i]\) for candidate classes \([T_1,\dots,T_N]\). Image and text representations are compared via cosine similarity, and class probabilities are computed as:
\begin{equation}
p(y=i|I)=\frac{\exp(\cos(f(I), g(E_w(T_i)))/\tau)}{\sum_{j=1}^N
\exp(\cos(f(I), g(E_w(T_j)))/ \tau)}  
\end{equation}
In this context, \(\tau\) is the temperature coefficient, and \(\cos(\cdot,\cdot)\) represents the cosine similarity between vision and language features.

\textbf{Inherent Calibration Capabilities of Pre-trained CLIP.}
CLIP learns robustly aligned image--text representations
through contrastive pre-training on massive-scale image-text pairs~\cite{CLIP,coop,promptsrc}. Prior studies~\cite{modern,closelook,promptsrc} indicate that the contrastive constraints during the pre-training phase allow the model to maintain a rational probability distribution when encountering open-vocabulary samples. This is because the original CLIP model, without being fine-tuned on downstream tasks, leverages the image-text alignment knowledge acquired during pre-training to provide predictive results that are consistent with its actual accuracy. In summary, large-scale pre-training provides the CLIP model with robust image-text alignment and strong calibration capabilities.

\section{Analysis}
\subsection{Logit Shift in Fine-Tuned CLIP}
Fine-tuning methods~\cite{cocoop,maple,promptsrc,lora}, such as prompt learning and LoRA, effectively improve a model's discriminative accuracy on specific downstream datasets, yielding accuracy improvements on train categories. However, as analyzed in~\cite{kgcoop,prograd,cocoop}, task-specific optimization during fine-tuning inevitably leads to the forgetting of general alignment knowledge acquired during pre-training. For instance, CoOp~\cite{coop}, which focuses solely on learning textual prompts, inherently disrupts the alignment between textual and visual representations. 

To mitigate confidence misalignment induced by fine-tuning, leveraging a well-aligned pre-trained model as an anchor can effectively constrain the generalization boundaries of downstream models. For instance, PromptSRC~\cite{promptsrc} utilizes features from the original CLIP for regularization, while PPO and GRPO~\cite{ppo,grpo} algorithms in reinforcement learning employ KL divergence to penalize excessive model updates. Here, we adopt the original CLIP, which retains its pre-trained vision-language alignment, as a calibration anchor. We define a shift-aware metric ($z$) that measures the sample-wise discrepancy between the logits of the original and fine-tuned CLIP models:

\begin{equation}
\label{eq:shift-score}
z = \frac{1}{N}\sum_{i=1}^{N}|P_{i}-\hat{P}_{i}|
\end{equation}

where $N$ denotes the number of classes, and $P$ and $\hat{P}$ are the logits produced by the fine-tuned and original CLIP, respectively. Thus, $z$ is explicitly a \emph{logit-shift score}, rather than a direct distance between hidden image or text representations. It summarizes both changes in the top-class score and changes across the full class vector, providing an observable prediction-space signal of how strongly fine-tuning moves a sample away from the original CLIP anchor. We use the $L_1$ distance because it responds linearly to coordinate-wise changes, is less dominated by a small number of large deviations than $L_2$, and avoids the Softmax and logarithmic operations required by KL divergence.

\textbf{Why is the original CLIP a useful anchor?} We do not assume that the original CLIP is perfectly calibrated or that its prediction should replace the fine-tuned output. Instead, it provides a fixed, label-free reference that retains the broad image--text alignment learned during pre-training. Fine-tuning is expected to change the logits when it learns useful task-specific information, so a nonzero $z$ is not itself an error. The relevant observation is that excessive departure from this reference can continue after accuracy has saturated and is then accompanied by increasing calibration error, as shown below. SAC therefore uses the \emph{magnitude} of the departure to control calibration strength, while preserving the class ordering produced by the fine-tuned model. This separates the roles of the two models: the fine-tuned CLIP determines the prediction, whereas the original CLIP only indicates how strongly its confidence should be adjusted.

\subsection{Analysis of Logit Shift}
\begin{figure}[t]
      \centering
      \begin{subfigure}{0.4\textwidth}
        \centering
        \includegraphics[width=\linewidth]{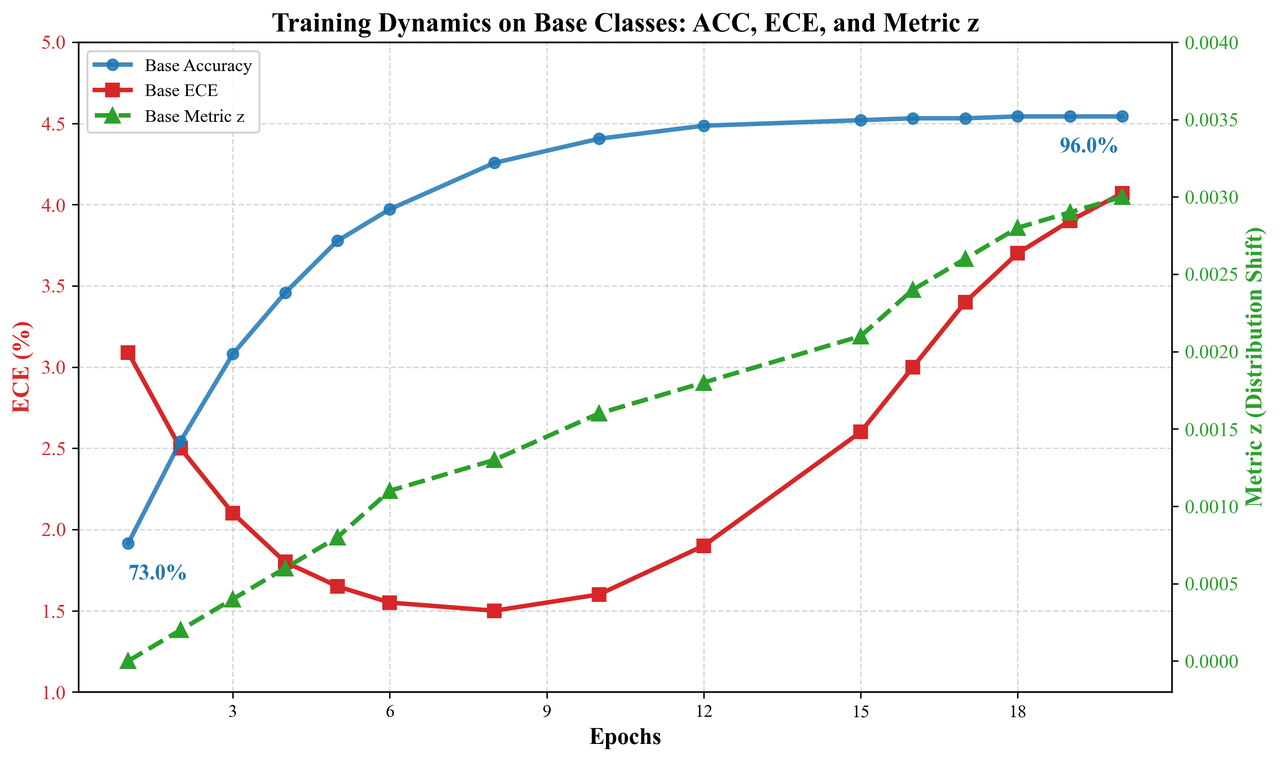}
        \caption{Performance on Train Classes}
        \label{fig:sub1}
      \end{subfigure}
      \begin{subfigure}{0.4\textwidth}
        \centering
        \includegraphics[width=\linewidth]{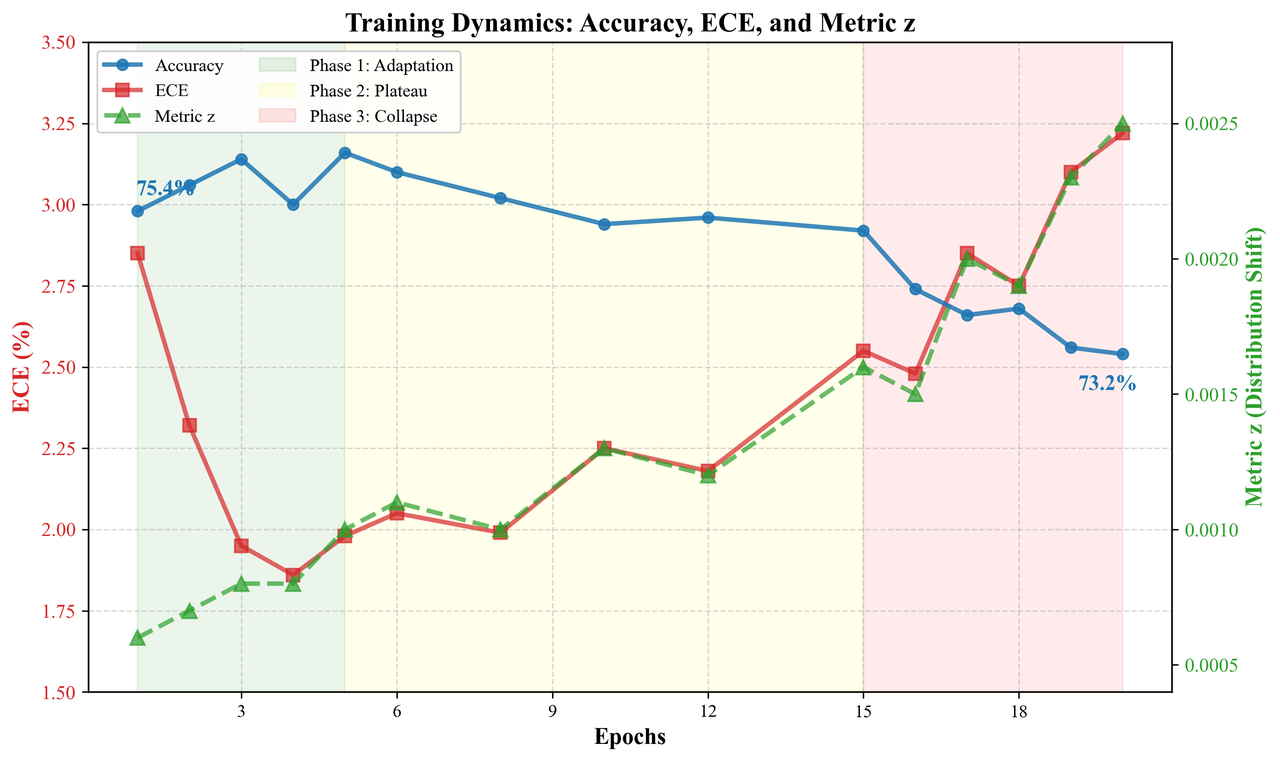}
        \caption{Performance on Unseen Classes}
        \label{fig:sub2}
      \end{subfigure}
      \caption{Evolution of MaPLe on Stanford Cars during fine-tuning. The logit-shift score $z$, accuracy, and ECE change together as training progresses on train and unseen classes.}
    \label{fig:SAC}
\end{figure}
To understand whether $z$ is useful for calibration, we follow the model throughout fine-tuning rather than inspecting only the final checkpoint. At every epoch, we measure accuracy, ECE, and the average logit-shift score on the train and unseen test sets (Figure~\ref{fig:SAC}). The original CLIP remains fixed and therefore acts as a common reference: a small $z$ means that the fine-tuned model still produces logits similar to the pre-trained model, whereas a large $z$ means that fine-tuning has substantially changed its output distribution. This comparison lets us examine whether moving farther from the original CLIP is accompanied by a change in calibration.

\begin{figure*}[t]
  \centering
  \begin{subfigure}{0.27\textwidth}
    \centering
    \includegraphics[width=\linewidth]{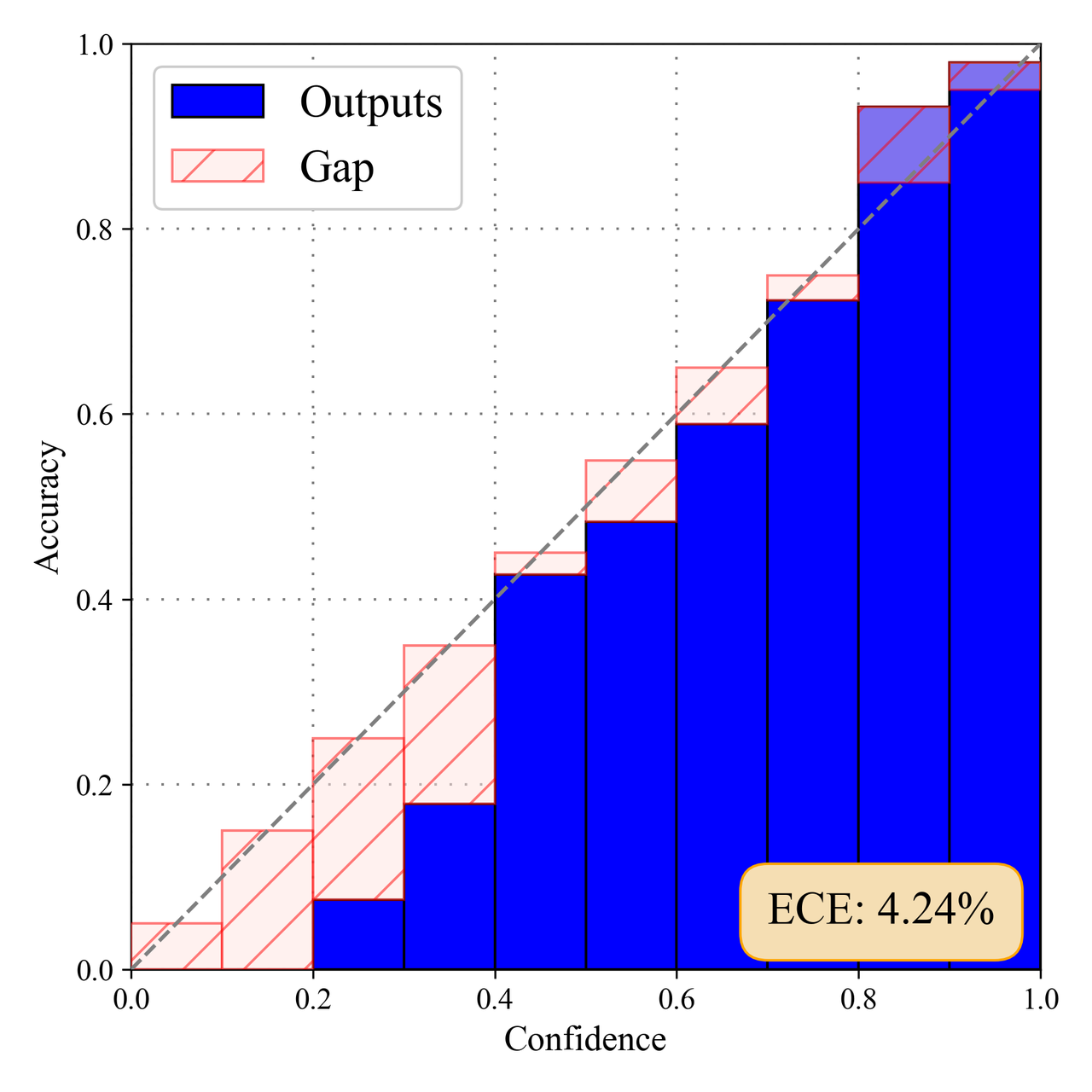}
    \caption{Well Calibration}
  \end{subfigure}
  \begin{subfigure}{0.27\textwidth}
    \centering
    \includegraphics[width=\linewidth]{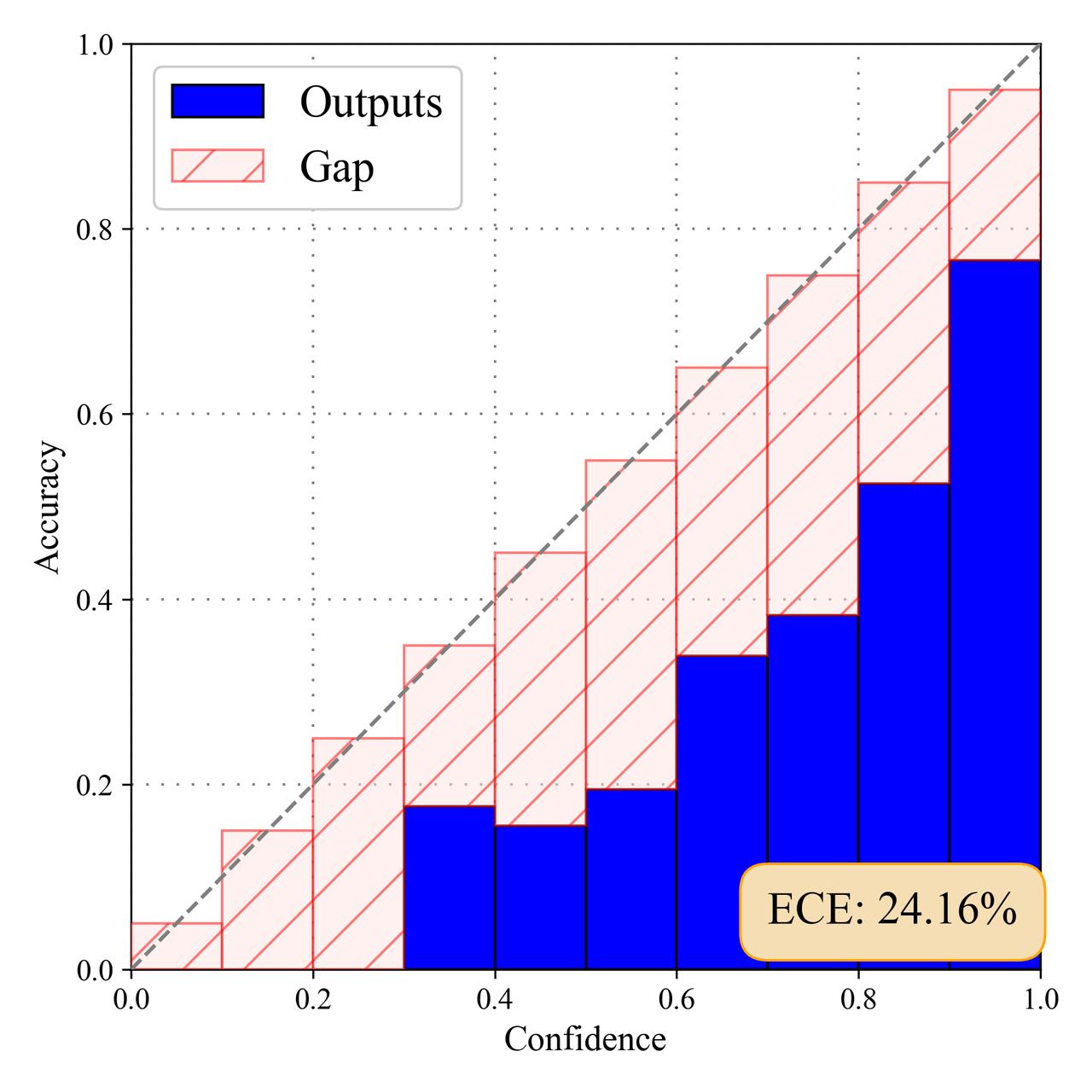}
    \caption{Over-confidence}
  \end{subfigure}
    \begin{subfigure}{0.27\textwidth}
    \centering
    \includegraphics[width=\linewidth]{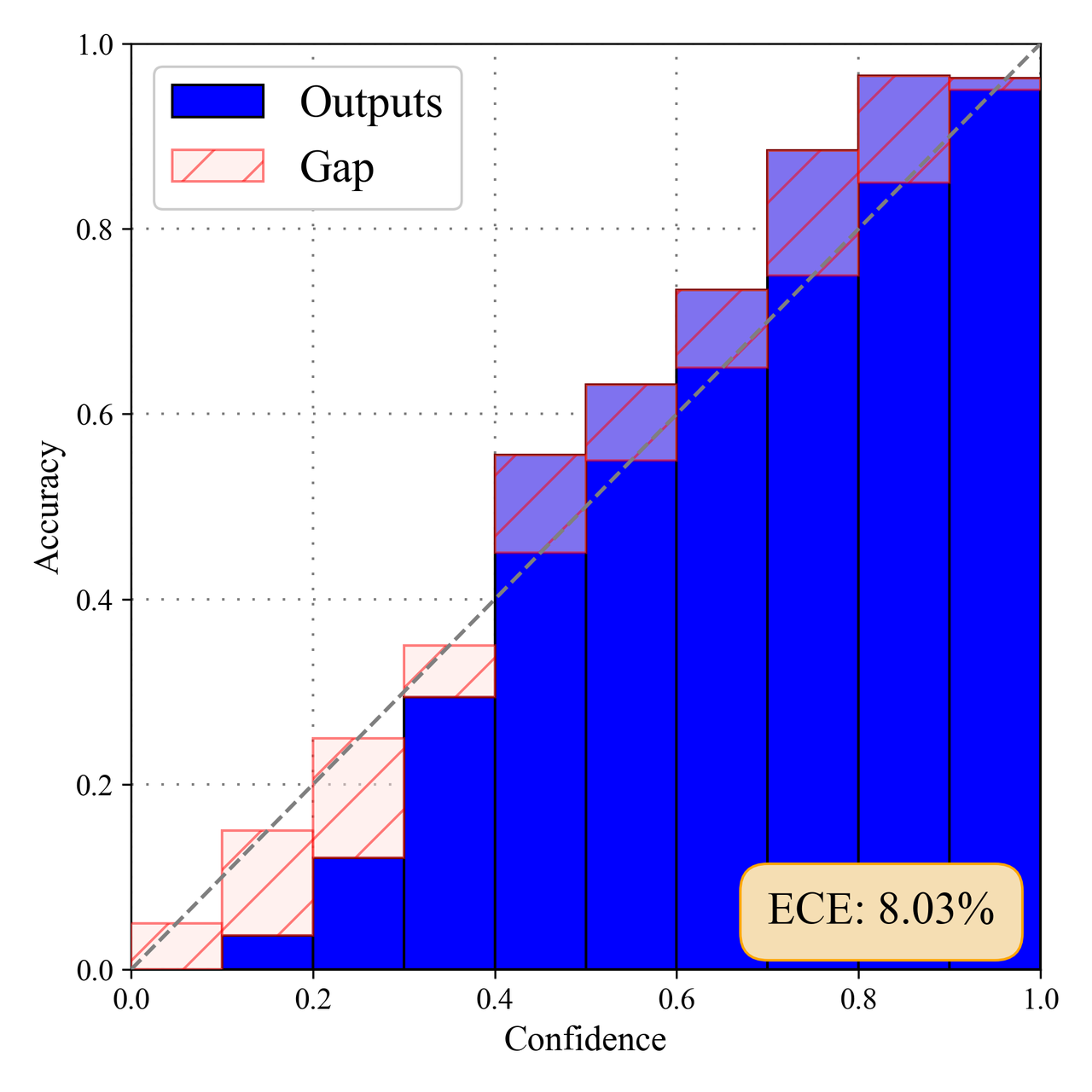}
    \caption{Under-confidence}
  \end{subfigure}
  \caption{Reliability diagrams of MaPLe on Stanford Cars at different training stages. Bars below the diagonal indicate overconfidence, while bars above it indicate underconfidence. Subfigures (a)--(c) show a relatively well-calibrated state, a strongly overconfident state, and an underconfident state, respectively.}
  \label{fig:ece}
\end{figure*}

\textbf{Early stage: useful adaptation.} At the beginning of training, accuracy improves quickly while $z$ remains small. ECE also decreases, especially on the unseen classes. This indicates that the model can learn task-relevant information without immediately losing the useful confidence structure inherited from CLIP. In this region, a strong correction is unnecessary because the fine-tuned and original models remain close.

\textbf{Middle stage: accuracy saturates before the logits stop moving.} On the train classes, accuracy approaches a plateau, but $z$ continues to grow. ECE first reaches its minimum and then increases. On the unseen classes, the same effect is clearer: accuracy changes only gradually, whereas $z$ and ECE rise together after the early optimum. In other words, continued fine-tuning changes the model's confidence distribution more than it improves classification. This explains why accuracy alone is insufficient for selecting a well-calibrated checkpoint and why the distance from the original CLIP can provide additional information.

\textbf{Late stage: confidence becomes increasingly misaligned.} In later epochs, the train-class accuracy remains nearly unchanged and the unseen-class accuracy declines, while both $z$ and ECE increase markedly. The model is therefore not simply becoming different from CLIP; the additional output change is accompanied by worse calibration. The reliability diagrams in Figure~\ref{fig:ece} provide an intuitive view of this phenomenon. Bars below the diagonal indicate that confidence exceeds accuracy, i.e., overconfidence, while bars above the diagonal indicate underconfidence. The examples show that fine-tuning can produce both cases, although overconfidence is more severe in the illustrated failure state.

These observations motivate SAC in a direct way. First, the correction should be sample-specific because different inputs can move different distances from the CLIP anchor. Second, a large $z$ should generally produce a smaller coefficient to soften the overconfident logits. Third, the coefficient must also be able to exceed $1$ to correct underconfident predictions. Thus, the training dynamics and reliability diagrams determine the desired behavior of the mapping in Equation~\eqref{eq:sac-weight}. This evidence is illustrative rather than causal, so we further test the relationship across methods and datasets using Mutual Information in the experiments.

\section{Shift-Aware Calibration}
According to our analysis, $z$ serves as an indicator of logit shift between the original and fine-tuned VLMs. However, this metric is insufficient to calibrate model outputs directly, as factors such as numerical variances and value range constraints must be addressed to ensure a valid transformation. Therefore, we design the following function to transform \(z\) into SAC weights:

\begin{equation}
\label{eq:sac-weight}
\gamma = \alpha \cdot e^{-kz}.
\end{equation}

Directly using $z$ as a multiplier would produce the wrong monotonic behavior: larger deviations would increase rather than suppress confidence. It would also restrict the coefficient to the numerical range of the raw distance and provide no explicit neutral point. We therefore map $z$ through $\alpha e^{-kz}$ according to the following principles:
\begin{itemize}
    \item \textbf{Negative-correlation mapping}: Overconfidence is the dominant failure mode when fine-tuning substantially changes the output vector. The exponential decay makes large $z$ produce a smaller coefficient, which softens the logits. When $z$ is small, the coefficient approaches $\alpha$ and can exceed $1$, allowing SAC to sharpen predictions for underconfident samples. The point $z=\ln(\alpha)/k$ gives $\gamma=1$ and therefore defines a neutral shift at which the original logits are preserved.
    \item \textbf{Bounded and smooth transformation}: The range $(0,\alpha]$ prevents extreme positive scaling and guarantees a positive coefficient. In contrast, $1/x$ and $-\ln x$ are singular near zero and can react excessively to small numerical differences between two closely aligned models. The exponential mapping changes smoothly over the full nonnegative domain, which is desirable because neighboring samples should not receive abruptly different calibration strengths.
    \item \textbf{Roles of $k$ and $\alpha$}: The normalized CLIP similarities often yield relatively small values of $z$. The parameter $k$ controls sensitivity to these differences: a larger $k$ produces faster decay and stronger separation among samples. The parameter $\alpha$ controls the maximum coefficient and the location of the neutral point. Their roles are therefore complementary rather than interchangeable, which is also reflected by the degradation observed when either component is removed in Table~\ref{tab-ablation-internal}.
\end{itemize}

\subsection{Optional SAC-Plus Refinement}
The base SAC mapping is the core method and already provides effective calibration with only $k$ and $\alpha$. For completeness, we also introduce SAC-Plus as an optional refinement for samples whose scaling coefficients are far from $1$. SAC-Plus leaves moderate coefficients unchanged and strengthens corrections at the two extremes:
\begin{equation}
\label{eq:sac-plus}
    \hat{\gamma} =
\begin{cases}
\gamma^2, & \text{if } \gamma < \lambda_1, \\
\gamma, & \text{if } \lambda_1 \leq \gamma \leq \lambda_2, \\
\gamma^2, & \text{if } \gamma > \lambda_2.
\end{cases}
\end{equation}
where $\hat{\gamma}$ is the SAC-Plus coefficient and $\lambda_1<1\leq\lambda_2$ define a no-change interval. Because all coefficients are positive, their effect can be read directly from their distance to $1$: $\gamma<1$ softens the logits and suppresses overconfidence, whereas $\gamma>1$ sharpens the logits and mitigates underconfidence. Squaring strengthens both corrections: for $0<\gamma<\lambda_1$, $\gamma^2<\gamma$ produces stronger softening; for $\gamma>\lambda_2$, $\gamma^2>\gamma$ produces stronger sharpening. Values in $[\lambda_1,\lambda_2]$ are retained to avoid unnecessary changes near the neutral scale. Importantly, $\lambda_1$ and $\lambda_2$ are fixed constants rather than learned or dataset-specific parameters; SAC-Plus introduces no additional training and is not required for the validity of the base SAC mechanism.

\subsection{Calibrated Inference}
Given an input image \(i\), we first collect the SAC scores of this image, denoted as \(\hat{\gamma}_i\), which is then used to calculate the rectified logits as follows:
\begin{equation}
P^{SAC}_i = \hat{\gamma}_i * \tau_{temp} * P_i
\end{equation}
where $P_i$ denotes the fine-tuned CLIP logit vector. Since $\hat{\gamma}_i\tau_{temp}>0$, the transformation preserves the ordering of class logits and therefore leaves the predicted class unchanged. SAC requires no additional training and applies a sample-wise confidence correction using the original CLIP anchor. The same fixed configuration is used across the main experiments; optional task-specific tuning can be performed when labeled validation data are available.

\begin{table*}[t]
    \centering
    \setlength{\tabcolsep}{2mm} 
    \begin{tabular}{lccc@{\hspace{3mm}}ccc@{\hspace{3mm}}ccc@{\hspace{3mm}}ccc}
    \toprule
    \multirow{2}{*}{\textbf{Method}} & \multicolumn{3}{c}{\textbf{ECE(\(\downarrow\))}} & \multicolumn{3}{c}{\textbf{ACE(\(\downarrow\))}} & \multicolumn{3}{c}{\textbf{MCE(\(\downarrow\))}} & \multicolumn{3}{c}{\textbf{PIECE(\(\downarrow\))}}  \\  
    \cmidrule(r){2-4} \cmidrule(r){5-7} \cmidrule(r){8-10} \cmidrule(r){11-13} 
        ~ & Conf & DAC & SAC & Conf & DAC & SAC & Conf & DAC & SAC & Conf & DAC & SAC \\ \midrule
        CoCoOp & 5.44 & 5.70 & \textbf{4.24} & 5.35 & 5.60 & \textbf{4.22} & 1.38 & 1.40 & \textbf{1.20} & 7.35 & 8.06 & \textbf{6.83} \\
        KgCoOp & 3.98 & 4.11 & \textbf{3.85} & 3.93 & 4.09 & \textbf{3.78} & \textbf{1.08} & 1.18 & 1.10 & 6.45 & 6.62 & \textbf{6.39} \\
        MaPLe & 7.80 & 5.91 & \textbf{5.35} & 7.77 & 5.93 & \textbf{5.30} & 2.08 & 1.62 & \textbf{1.61} & 9.53 & 8.19 & \textbf{7.69} \\
        ProGrad & 5.04 & 6.13 & \textbf{4.04} & 4.95 & 6.18 & \textbf{4.05} & 1.47 & 1.53 & \textbf{1.24} & 7.41 & 8.20 & \textbf{6.75} \\
        PromptSRC & 4.29 & 4.55 & \textbf{3.47} & 4.24 & 4.41 & \textbf{3.40} & 1.16 & 1.17 & \textbf{1.03} & 6.70 & 6.82 & \textbf{6.12} \\
    \bottomrule
    \end{tabular}
    \caption{Average calibration performance across 11 datasets on unseen classes. We report the SAC-Plus results. Detailed results, including standard deviations, are provided in the supplementary material.}
    \label{tab-test}
\end{table*}

\begin{table*}[t]
    \centering
    \small
    \setlength{\tabcolsep}{1mm}
    \begin{tabular}{lcccccccc}
    \toprule
    \multirow{2}{*}{\textbf{Method}} 
        & \multicolumn{4}{c}{\textbf{Cross-Dataset}} 
        & \multicolumn{4}{c}{\textbf{Domain-Generalization}} \\
    \cmidrule(lr){2-5} \cmidrule(lr){6-9}
        & ECE (Conf) & ECE (SAC) & ACE (Conf) & ACE (SAC) 
        & ECE (Conf) & ECE (SAC) & ACE (Conf) & ACE (SAC) \\
    \midrule
    CoCoOp    & 4.28 & \textbf{4.15} & 4.39 & \textbf{4.15} & 3.97 & \textbf{3.37} & 3.96 & \textbf{3.34} \\
    KgCoOp    & 4.37 & \textbf{4.15} & 4.33 & \textbf{4.11} & 4.49 & \textbf{2.96} & 4.53 & \textbf{3.00} \\
    MaPLe     & 4.04 & \textbf{3.72} & 4.07 & \textbf{3.76} & 4.10 & \textbf{2.94} & 4.18 & \textbf{3.01} \\
    ProGrad   & 4.16 & \textbf{4.03} & 4.06 & \textbf{4.04} & 4.44 & \textbf{3.18} & 4.39 & \textbf{3.17} \\
    PromptSRC & 4.40 & \textbf{3.92} & 4.33 & \textbf{3.93} & 4.10 & \textbf{3.89} & 4.09 & \textbf{3.94} \\
    \bottomrule
    \end{tabular}
    \caption{Average calibration performance under cross-dataset and domain-generalization settings.}
    \label{tab:ece_ace_combined}
\end{table*}

\section{Experiments}
\subsection{Experimental Setup}
\label{exp_set}
\textbf{Evaluation Paradigm.} Following standard open-vocabulary evaluation~\cite{maple,cocoop}, each dataset is divided into train and unseen classes. The model is tuned with few-shot data from the train classes, and we report calibration on both groups. We additionally evaluate cross-dataset and domain-generalization settings, for which methods fitted to the source distribution and the original formulation of DAC are not directly applicable.

\textbf{Compared Methods.} We mainly focus on benchmarking against the other 5 current representative prompt learning methods: CoCoOp~\cite{cocoop}, KgCoOp~\cite{kgcoop}, MaPLe~\cite{maple}, ProGrad~\cite{prograd}, and PromptSRC~\cite{promptsrc}. Since models like CoOp only consider the text modality and exhibit low accuracy, making their calibration significance minimal, we focused on testing and calibrating its optimized version, CoCoOp and KgCoOp. For train classes calibration, we select three representative calibration methods: Histogram Binning (HB)~\cite{bin1}, Isotonic Regression (IR)~\cite{bin2}, Multi-Isotonic Regression (MIR)~\cite{bin3}, and Temperature Scaling (TS)~\cite{ts}. For unseen classes calibration, we compare the SOTA method designed for CLIP, DAC~\cite{DAC}. \textbf{Datasets} and \textbf{Evaluation Metrics} used in the experiments are shown in Appendix.

\textbf{Implementation details.} For three settings, we use CLIP (ViT-B/16) as the pre-trained VLM throughout our experiments and report results averaged over 3 runs. SAC is the primary method; SAC-Plus is reported as an optional fixed-rule refinement to quantify the additional gain obtainable without learning extra parameters. 
\textbf{To ensure a fair comparison across test distributions, we use one fixed configuration rather than per-dataset tuning.} The base SAC method uses only $k=15$ and $\alpha=1.10$. For the optional SAC-Plus refinement, we fix $\lambda_1=0.9$ and $\lambda_2=1.0$ globally; these thresholds are not learned and are not tuned per dataset. This protocol evaluates transfer of the calibration rule without using target-dataset labels for parameter selection. Additional implementation and sensitivity details are provided in the supplementary material.

\subsection{Empirical Results}
\subsubsection{RQ1:} How Does SAC Compare with Existing Calibration Methods across Open-World Settings?

DAC derives calibration weights from distances between textual representations of train and unseen classes, which is designed primarily for the within-dataset open-vocabulary setting. Because these weights do not incorporate sample-level visual shifts, the original formulation is not directly applicable to the cross-dataset and domain-generalization settings considered here.

\paragraph{Unseen-class calibration.} Table~\ref{tab-test} evaluates five fine-tuning methods with four complementary calibration metrics. SAC obtains lower ECE and ACE than both the uncalibrated model and DAC in all five rows. Averaged over the five methods, ECE decreases from 5.31 to 4.19 (a 21.1\% relative reduction), while ACE decreases from 5.25 to 4.15. The improvement is not limited to a single adaptation strategy: ECE decreases from 5.44 to 4.24 for CoCoOp, from 3.98 to 3.85 for KgCoOp, from 7.80 to 5.35 for MaPLe, from 5.04 to 4.04 for ProGrad, and from 4.29 to 3.47 for PromptSRC. This breadth matters because the five methods regularize CLIP in different ways; a calibration rule tied to one prompt-learning objective would not be expected to improve all of them with the same parameters.

The gains also hold across metrics with different emphases. ECE and ACE evaluate average calibration under fixed and adaptive partitions, MCE emphasizes the largest bin-wise error, and PIECE incorporates neighborhood information. SAC gives the best post-hoc result in every reported method--metric pair. In contrast, DAC increases ECE for CoCoOp, KgCoOp, ProGrad, and PromptSRC, while helping MaPLe. This pattern indicates that a class-level textual prior can be useful for some representation changes but is not uniformly aligned with sample-specific confidence errors. SAC instead conditions the correction on each input's deviation from the original CLIP, which provides a direct explanation for its more consistent behavior.

\paragraph{Transfer across distribution shifts.} Table~\ref{tab:ece_ace_combined} tests whether the same fixed rule remains effective outside the base-to-novel setting. In cross-dataset evaluation, the average ECE across methods decreases from 4.25 to 3.99, and SAC improves every row, including 4.04 to 3.72 for MaPLe and 4.40 to 3.92 for PromptSRC. The improvements are larger under domain generalization: the average ECE decreases from 4.22 to 3.27 (a 22.6\% relative reduction), with reductions from 4.49 to 2.96 for KgCoOp, from 4.10 to 2.94 for MaPLe, and from 4.44 to 3.18 for ProGrad. ACE follows the same overall pattern. Because no target-specific calibration parameters are fitted, these results show that the original CLIP anchor remains informative when either the class vocabulary or the visual domain changes. Dataset-level results in the supplementary material further show that the averages are not produced by a single target dataset.

\begin{table}[t]
    \centering
    \setlength{\tabcolsep}{1mm}
    \begin{tabular}{ccccccc}
    \toprule
    \multirow{2}{*}{\textbf{Method}} & \multirow{2}{*}{\textbf{Conf}} & \multicolumn{4}{c}{\textbf{Training-free}} & \textbf{Training} \\ 
    \cmidrule(l){3-6} \cmidrule(l){7-7} 
                                     &                                 & \textbf{IR} & \textbf{HB} & \textbf{MIR} & \textbf{SAC} & \textbf{TS} \\ 
    \midrule
    CoCoOp & 3.60 & 7.80 & 7.50 & 3.87 & \textbf{3.05} & 3.42 \\
    KgCoOp & 5.87 & 7.23 & 7.41 & 7.38 & 4.10 & \textbf{3.01} \\
    MaPLe & 2.80 & 7.81 & 6.63 & 2.70 & \textbf{2.50} & 2.64 \\
    ProGrad & 5.93 & 5.69 & 5.63 & 4.42 & 4.17 & \textbf{3.06} \\
    PromptSRC & 3.74 & 6.39 & 6.38 & 3.55 & \textbf{2.75} & 2.86 \\
    \bottomrule
    \end{tabular}
    \caption{Average ECE results across 11 datasets on train classes. Smaller values are better.}
    \label{tab-train}
\end{table}

\begin{table*}[t]
    \centering
    \begin{tabular}{>{\centering\arraybackslash}p{1.5cm}
                    >{\centering\arraybackslash}p{1cm}
                    >{\centering\arraybackslash}p{1cm}
                    >{\centering\arraybackslash}p{1cm}
                    >{\centering\arraybackslash}p{1.5cm}
                    >{\centering\arraybackslash}p{1.5cm}
                    >{\centering\arraybackslash}p{1.5cm} 
                    >{\centering\arraybackslash}p{1cm}
                    >{\centering\arraybackslash}p{2cm}
                    >{\centering\arraybackslash}p{1cm}
                    }
    \toprule
    \textbf{Method} & \textbf{Conf} & \textbf{w/o \(k\)} & \textbf{w/o \(\alpha\)} & \textbf{w/ \(L_2\)} & \textbf{w/ 1/x} & \textbf{w/ -ln(x)} & \textbf{DAC} & \textbf{Base SAC} & \textbf{SAC} \\ \midrule
    CoCoOp    & 5.44 & 9.08  & 9.73 & 5.65 & 11.15 & 15.68 & 5.70 & \underline{4.72} & \textbf{4.24} \\ 
    KgCoOp    & 3.98 & 7.52  & 8.12 &  4.34 & 26.74 & 27.89  & 4.11 & \underline{3.97} & \textbf{3.85} \\ 
    MaPLe     & 7.80 & 11.32 & 11.25 & 6.50 & 28.13 & 21.76 & \underline{5.91} & 6.57 & \textbf{5.35} \\ 
    ProGrad   & 5.04 & 8.92  & 11.84 & 4.89 & 28.85 & 24.34 & 6.13 & \underline{4.44} & \textbf{4.04} \\ 
    PromptSRC & 4.29 & 7.80  & 9.41 & 4.63 & 23.87 & 19.56 & 5.28 & \underline{4.71} & \textbf{3.47} \\ \bottomrule
    \end{tabular}
    \caption{Ablation study of the core SAC components and the optional SAC-Plus refinement.}
    \label{tab-ablation-internal}
\end{table*}


\begin{table}[t]
\centering
\setlength{\tabcolsep}{1mm}
\begin{tabular}{lcccc|c}
\toprule
\textbf{Method} & \textbf{DTD} & \textbf{Cars} & \textbf{SUN397} & \textbf{Food101} & \textbf{Average} \\ \midrule
CoCoOp    & 0.737 & 0.612 & 0.584 & 0.605 & \textbf{0.635} \\
KgCoOp    & 0.812 & 0.658 & 0.741 & 0.692 & \textbf{0.726} \\
ProGrad   & 0.481 & 0.339 & 0.412 & 0.455 & \textbf{0.422} \\
MaPLe     & 0.887 & 0.529 & 0.697 & 0.754 & \textbf{0.717} \\
PromptSRC & 0.782 & 0.666 & 0.594 & 0.567 & \textbf{0.652} \\ 
\bottomrule
\end{tabular}
\caption{Mutual Information (MI) between logit-shift score $z$ and ECE across four representative datasets. The supplementary material extends the analysis to other datasets.}
\label{tab:mi_results}
\end{table}

\subsubsection{RQ2:} Compared with traditional methods, can SAC achieve performance improvements on the train classes?

As previously analyzed, while fine-tuning generally enhances model accuracy, it consistently introduces confidence misalignment—predominantly over-confidence, alongside occasional under-confidence. By leveraging \textbf{the degree of distribution shift} to quantify such misalignment, SAC seamlessly generalizes to the calibration of seen classes, further demonstrating the inherent versatility of its design. We compared SAC with two representative types of calibration methods in Table~\ref{tab-train}: 
\begin{itemize} 
    \item Compared with traditional training-free calibration methods, such as Bin-based methods (IR, HB, MIR), SAC consistently outperforms all of them, demonstrating its strong ability to calibrate CLIP. Notably, these methods often exhibit significant performance fluctuations across different fine-tuning methods, sometimes even increasing ECE rather than reducing it as intended. The above results demonstrate that binning-based calibration cannot effectively handle subtle biases introduced by fine-tuning; instead, it yields counterproductive outcomes. 
    \item Compared with the training-based method (TS), which learns calibration coefficients tailored to individual datasets it exhibits superior performance under certain prompt learning methods. Nevertheless, training-free SAC achieves performance comparable to trained methods and demonstrates its plug-and-play advantage.
\end{itemize}

\subsubsection{RQ3:}  Does the SAC improve the calibration performance, as designed?

\paragraph{The Effect of Each Component in SAC:}
\begin{itemize}
    \item \textbf{Without \(\alpha\) / \(k\)}: Removing \(\alpha\) makes SAC fail to handle underconfident datasets and yields degraded performance, while ablating \(k\) impairs the model’s capacity to distinguish representation shift differences, consistent with our original design motivation. This demonstrates that both \(\alpha\) (which expands the value range) and \(k\)-based amplification are indispensable for optimal performance.
    \item \textbf{With other distance}: We compared the effects of $L_1$ and $L_2$ distances in measuring the distribution shift $z$. The above analysis shows no significant difference in different distance. Given its lower computational complexity and higher efficiency in high-dimensional spaces, we adopt the $L_1$ distance as the default setting. 
    \item \textbf{With other function}: We conducted experiments under the same settings by replacing the \(e^{-x}\) function with \(1/x\) and \(-\ln(x)\) function. Since the values of these functions fluctuate drastically with respect to \(x\), they fail to handle confidence imbalances.
    
\end{itemize}

\paragraph{Base SAC and the optional SAC-Plus refinement.} The ``SAC'' column evaluates the core method using only $k$ and $\alpha$. As shown in Table~\ref{tab-ablation-internal}, base SAC improves calibration for most fine-tuning methods, demonstrating that the original CLIP anchor and the logit-shift score already form an effective calibration mechanism. SAC-Plus further improves all rows by strengthening corrections for coefficients outside the neutral interval. However, it is only an optional deterministic refinement: $\lambda_1$ and $\lambda_2$ are fixed globally, introduce no learned parameters, and are not tuned per dataset. Therefore, SAC is the main methodological contribution, while SAC-Plus illustrates that the core method can be further enhanced with a simple fixed rule.

\paragraph{Information dependency between $z$ and ECE.}
We estimate MI between the mean logit-shift score and ECE along the
training trajectory using
\texttt{sklearn.mutual\_info\_regression}. As shown in Table~\ref{tab:mi_results}, all reported method--dataset
pairs yield positive MI. KgCoOp achieves the highest average MI
(0.726), while ProGrad has the lowest (0.422), consistent with its
gradient constraint limiting output drift. These results show that
the relationship between $z$ and ECE persists across datasets and
fine-tuning methods. Since SAC directly uses $z$ to determine its
sample-wise scaling coefficient, the MI analysis verifies that SAC
relies on calibration-relevant information rather than an arbitrary
distance. Together with the base-SAC improvements, this supports the
effectiveness of the core SAC mechanism.

\subsubsection{RQ4:} To handle complex downstream datasets, how should we adjust the parameters of SAC?

To address more challenging real-world scenarios, we provide a detailed analysis of the optimal parameter choices for each module in Appendix. Based on this analysis, we summarize the following parameter selection guidelines for SAC: for all baselines and datasets in the open-vocabulary setting, the default parameters \((\alpha = 1.10, k = 15)\) are generally sufficient, reflecting the parameter insensitivity of SAC. For different baselines or evaluation settings, \(\alpha\) can be adjusted based on reliability diagrams—specifically, increasing \(\alpha\) for underconfident models and decreasing it for overconfident ones, while \(k\) primarily scales the input magnitude. Overall, SAC effectively integrates our research findings into a unified confidence calibration framework, with extensive experiments confirming its parameter robustness and adaptability.

\section{Conclusion}
Prompt learning enables effective open-vocabulary adaptation but can introduce confidence misalignment. We show that fine-tuning-induced logit shift is statistically associated with calibration error and use this signal to construct our method SAC. SAC applies a positive, sample-wise logit scaling coefficient derived from the original CLIP anchor, preserving the predicted class without requiring target labels in the default protocol. Across train and unseen classes as well as cross-dataset and domain-generalization settings, SAC improves calibration for multiple fine-tuning methods. MI analysis verifies that the shift signal used by SAC is informative about ECE, supporting the proposed calibration mechanism.

\bibliography{aaai2027}


\end{document}